\documentclass[10pt,twocolumn,letterpaper]{article}

\usepackage{cvpr}
\usepackage{times}
\usepackage{epsfig}
\usepackage{graphicx}
\usepackage{amsmath}
\usepackage{amssymb}
\usepackage{booktabs}

\usepackage[breaklinks=true,bookmarks=false]{hyperref}

\cvprfinalcopy 

\def\cvprPaperID{****} 

\begin{document}

\title{The 3rd Person in Context (PIC) Workshop and Challenge at CVPR 2021:\\
Short-video Face Parsing Track\\
\rm {Technical Report:\\
Edge Aware Network for Face Parsing}
}

\author{Xiao Liu, Xiaofei Si, Jiangtao Xie\\
	Dalian University of Technology\\
	{\tt\small \{birdylx, sxf, jiangtaoxie\}@mail.dlut.edu.cn}
}

\maketitle

\begin{abstract}
   This is a short technical report introducing the solution of Team Rat for Short-video Parsing Face Parsing Track of The 3rd Person in Context (PIC) Workshop and Challenge at CVPR 2021.
   In this report, we propose an Edge-Aware Network (EANet) that uses edge information to refine the segmentation edge. To further obtain the finer edge results, we introduce edge attention loss that only compute cross entropy on the edges, it can effectively reduce the classification error around edge and get more smooth boundary. Benefiting from the edge information and edge attention loss, the proposed EANet achieves 86.16\% accuracy in the Short-video Face Parsing track of the 3rd Person in Context (PIC) Workshop and Challenge, ranked the third place.
\end{abstract}

\section{Introduction}
Face Parsing is a particular task in semantic segmentation, it needs to predict the semantic category of each pixel on the human face, such as hair, eye, mouth, nose, \emph{etc}. Compared with face detection and facial keypoint detection, the face parsing can provide more attribute information for further application. 

There are a lot of works~\cite{2020Object, li2020improving, chen2018encoderdecoder, liu2021swin, wu2020ainnoseg, long2015fully,wang2020solov2, he2018mask, tao2020hierarchical, lin2019face,ruan2018devil} studied face parsing and scene parsing. But the some of them ignore the edge information, mean while, in this challenge, there are accuracy requirements for edge segmentation. Motivated by this, we propose Edge-Aware Network (EANet) mining the edge information to improve segmentation results. 

\begin{table} 
\caption{The rank results of Short-video Face Parsing}
\begin{center}
\begin{tabular}{ccc}
\hline
Team & Rank & Accuracy \\
\hline\hline
TCParser & 1st & 86.95\\  
BUPT-CASIA & 2nd & 86.84\\    
rat(ours) & 3rd & 86.16\\
\hline
\end{tabular}
\end{center}
\end{table}

\section{Method}

We analyzed the data set and found that the head area of some people occupies a small proportion in the whole image. If we segment the entire image directly, a lot of background noise will be introduced. Therefore, we first extract the required ROI area, and then perform segmentation on the ROI area. We adopted a two-stage strategy that is detection first, then segmentation. Figure 1 shows our overall parsing pipeline.

\begin{figure*}[tb]
	\centering
	\begin{minipage}[b]{1.0\linewidth}
		\centering
		\includegraphics[width=1.0\textwidth]{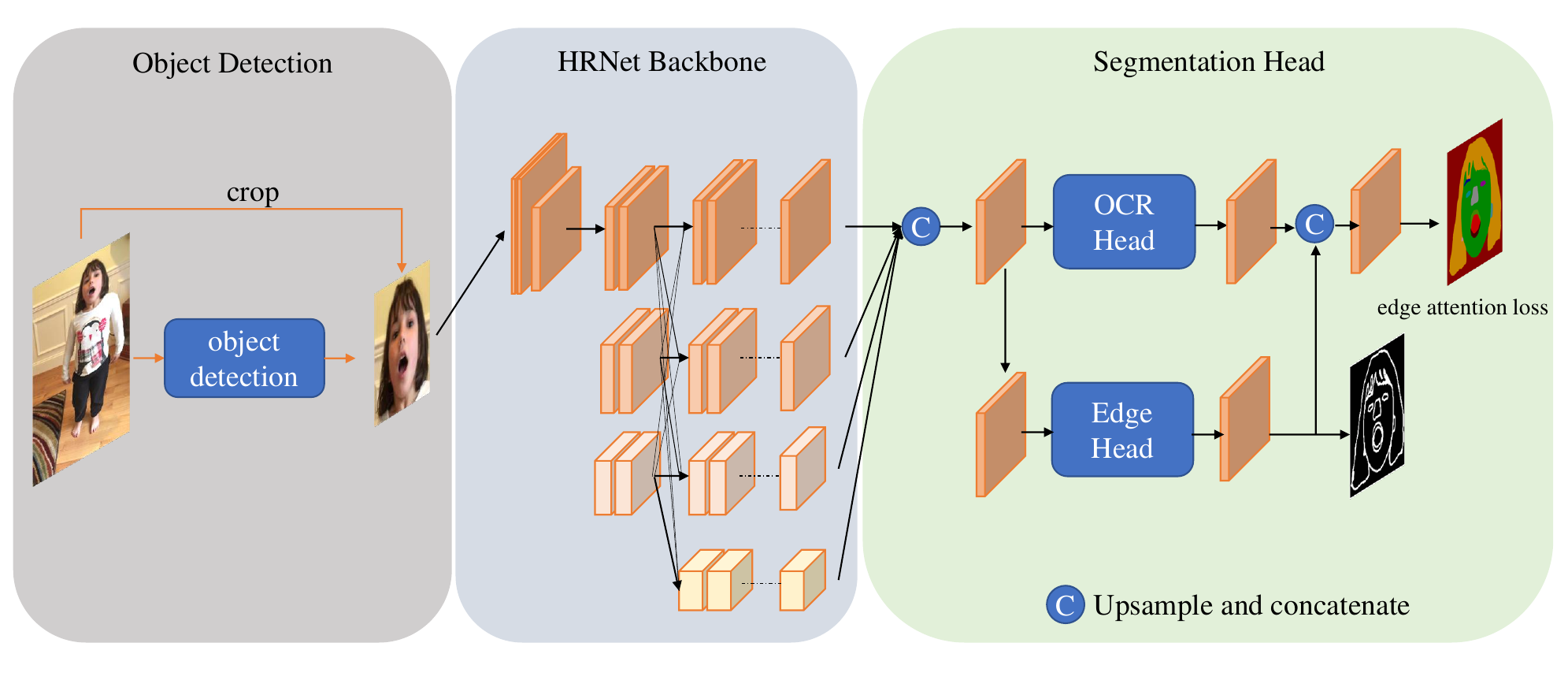}
	\end{minipage}
	\caption{Overview of our proposed scheme.}
	\label{figure:2}	
\end{figure*}

\subsection{ROI Detection}

We constructed a head region dataset based on the ground truth bounding box of the provided segmentation dataset for ROI region object detection. We trained a cascade rcnn~\cite{cai2019cascade} with ResNet50~\cite{he2016deep} backbone using FPN~\cite{2017Feature} and DCN~\cite{2020Deformable} to detect the head region. Since we need to roughly detect the ROI area, we do not need to be particularly accurate, but we need to detect all objects as much as possible.

\subsection{Face Segmentation}

We take OCRNet~\cite{2020Object} with HRNet~\cite{sun2019highresolution} backbone as our baseline, HRNet is a high-resolution network, it can capture detailed features well and avoid the loss of details due to continuous upsampling like other segmentation networks. We also need to consider the accuracy of the overall segmentation and the accuracy of the edge at the same time, so we designed a parallel edge segmentation branch on the basis of OCRNet, using edge supervision information to improve the accuracy of the segmentation result on the edge. 

As Figure 1 shows, we send the cropped ROI into the HRNet backbone, and then upsample and concatenate the four resolution feature maps to obtain the soft region. The soft region is then passed through the OCR branch and the Edge branch to get seg feature and edge feature, then concatenate and fusion edge with seg features to obtain the final segmentation result. In addition, we also adopt EdgeAttentionLoss~\cite{li2020improving} to use the obtained edge result as a mask, and only calculate the loss in the mask area to further strengthen the accuracy of the edge segmentation result.

\section{Experiments}

\subsection{Implementation details}
\noindent
\textbf{Datasets}. The dataset is selected from 1890 videos in the database. Each video clip takes 20 frames. There are 1490 videos in the trainingset, 200 videos in the validation set and 200 videos in the test set. Davis~\cite{Perazzi2016} J/F scores were used in final evaluation.

\noindent
\textbf{Training}. For face segmentation, the model input size is 448, due to the symmetry of the face, the horizontal flip data augmentation will introduce ambiguity, so we flip back those symmetry part of face like eyes, eyebrow, and we random rotate image with 90, 270 degree.There are some images with incomplete faces, so we use a random cut half augmentation to cut half of face. We trained the segmentation model with batch size of 6, SGD optimizer with poly~\cite{hu2019squeezeandexcitation} learning annealing, in total 40000 iterations. 

\noindent
\textbf{Testing}. We didn't use test time augmentation, we just simply enlarge the input size from 448 to 480.

\subsection{Experimental Results}
As shown in Table 2, proposed EANet improves 1.5 point compared with baseline OCRNet, we also discuss the add and concatenate operation in feature fusion stage. We ensemble baseline, EANet with add and concatenate fusion operation, total 3 models. Additional, as Figure 2 shows, we take grabcut~\cite{2004GrabCut} to further refine the missing part. Finally, our proposed  method achieves accuracy score 86.16\% on the challenge test set and ranks the 3rd place.

\begin{table} 
\caption{The results of our models on test set}
\begin{center}
\begin{tabular}{ccc}
\hline
Model & Backbone & Accuracy \\
\hline\hline
OCRNet & HRNet\_W48 & 83.42\\  
EANet & HRNet\_W48 & 84.94\\    
Ensemble & - & 85.92\\
Ensemble + grabcut & - & \textbf{86.16}\\
\hline
\end{tabular}
\end{center}
\end{table}

\section{Conlusion}
In this report, we propose an Edge-Aware Network mining the edge information to refine segmentation results, we adopt edge attention loss to further strengthen the edge to get more smooth edge and better results. The proposed method achieves 86.16\% accuracy and ranked the 3rd place.

\begin{figure}[tb]
	\centering
	\begin{minipage}[b]{1.0\linewidth}
		\centering
		\includegraphics[width=1.0\textwidth]{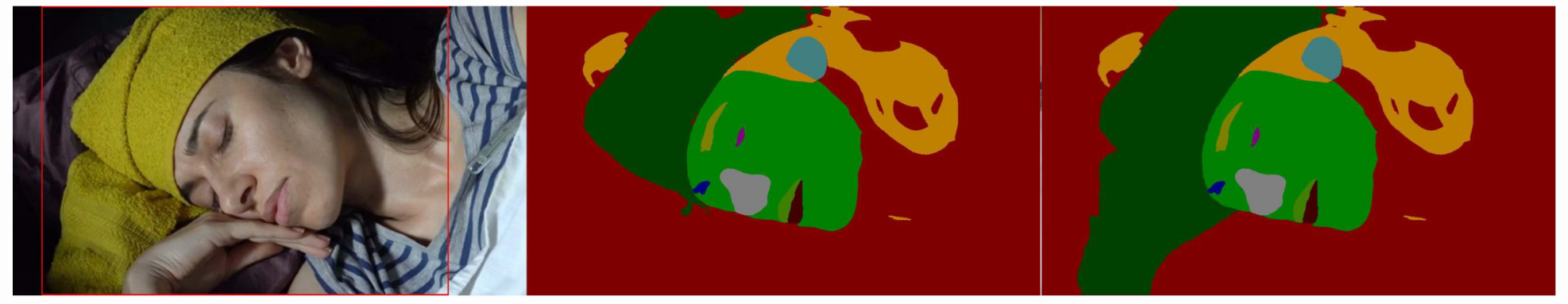}
	\end{minipage}
	\caption{The visualization result of the missing part of hat can be parsed after grabcut post process.}
	\label{figure:3}	
\end{figure}

\newpage
{\small
\bibliographystyle{ieee}
\bibliography{egbib}

\begin{thebibliography}{10}\itemsep=-1pt

\bibitem{cai2019cascade}
Z.~Cai and N.~Vasconcelos.
\newblock Cascade r-cnn: High quality object detection and instance
  segmentation, 2019.

\bibitem{chen2018encoderdecoder}
L.-C. Chen, Y.~Zhu, G.~Papandreou, F.~Schroff, and H.~Adam.
\newblock Encoder-decoder with atrous separable convolution for semantic image
  segmentation, 2018.

\bibitem{he2018mask}
K.~He, G.~Gkioxari, P.~Dollár, and R.~Girshick.
\newblock Mask r-cnn, 2018.

\bibitem{he2016deep}
K.~He, X.~Zhang, S.~Ren, and J.~Sun.
\newblock Deep residual learning for image recognition.
\newblock In {\em CVPR}, pages 770--778, 2016.

\bibitem{hu2019squeezeandexcitation}
J.~Hu, L.~Shen, S.~Albanie, G.~Sun, and E.~Wu.
\newblock Squeeze-and-excitation networks, 2019.

\bibitem{li2020improving}
X.~Li, X.~Li, L.~Zhang, G.~Cheng, J.~Shi, Z.~Lin, S.~Tan, and Y.~Tong.
\newblock Improving semantic segmentation via decoupled body and edge
  supervision, 2020.

\bibitem{lin2019face}
J.~Lin, H.~Yang, D.~Chen, M.~Zeng, F.~Wen, and L.~Yuan.
\newblock Face parsing with roi tanh-warping, 2019.

\bibitem{2017Feature}
T.~Y. Lin, P.~Dollar, R.~Girshick, K.~He, B.~Hariharan, and S.~Belongie.
\newblock Feature pyramid networks for object detection.
\newblock In {\em 2017 IEEE Conference on Computer Vision and Pattern
  Recognition (CVPR)}, 2017.

\bibitem{liu2021swin}
Z.~Liu, Y.~Lin, Y.~Cao, H.~Hu, Y.~Wei, Z.~Zhang, S.~Lin, and B.~Guo.
\newblock Swin transformer: Hierarchical vision transformer using shifted
  windows, 2021.

\bibitem{long2015fully}
J.~Long, E.~Shelhamer, and T.~Darrell.
\newblock Fully convolutional networks for semantic segmentation, 2015.

\bibitem{Perazzi2016}
F.~Perazzi, J.~Pont-Tuset, B.~McWilliams, L.~{Van Gool}, M.~Gross, and
  A.~Sorkine-Hornung.
\newblock A benchmark dataset and evaluation methodology for video object
  segmentation.
\newblock In {\em Computer Vision and Pattern Recognition}, 2016.

\bibitem{2004GrabCut}
C.~Rother.
\newblock Grabcut : Interactive foreground extraction using iterated graph
  cuts.
\newblock {\em Proceedings of Siggraph}, 23, 2004.

\bibitem{ruan2018devil}
T.~Ruan, T.~Liu, Z.~Huang, Y.~Wei, S.~Wei, Y.~Zhao, and T.~Huang.
\newblock Devil in the details: Towards accurate single and multiple human
  parsing, 2018.

\bibitem{sun2019highresolution}
K.~Sun, Y.~Zhao, B.~Jiang, T.~Cheng, B.~Xiao, D.~Liu, Y.~Mu, X.~Wang, W.~Liu,
  and J.~Wang.
\newblock High-resolution representations for labeling pixels and regions,
  2019.

\bibitem{tao2020hierarchical}
A.~Tao, K.~Sapra, and B.~Catanzaro.
\newblock Hierarchical multi-scale attention for semantic segmentation, 2020.

\bibitem{wang2020solov2}
X.~Wang, R.~Zhang, T.~Kong, L.~Li, and C.~Shen.
\newblock Solov2: Dynamic and fast instance segmentation, 2020.

\bibitem{wu2020ainnoseg}
J.~Wu, J.~Lu, X.~Kang, Y.~Zhang, Y.~Tang, J.~Song, Z.~Huang, S.~Ben, J.~Huang,
  and F.~Zhang.
\newblock Ainnoseg: Panoramic segmentation with high perfomance, 2020.

\bibitem{2020Object}
Y.~Yuan, X.~Chen, and J.~Wang.
\newblock Object-contextual representations for semantic segmentation.
\newblock In {\em European Conference on Computer Vision}, 2020.

\bibitem{2020Deformable}
X.~Zhu, H.~Hu, S.~Lin, and J.~Dai.
\newblock Deformable convnets v2: More deformable, better results.
\newblock {\em IEEE}, 2020.

\end{thebibliography}
}

\end{document}